\documentclass{article}
\usepackage{spconf,amsmath,graphicx}

\usepackage{color}
\usepackage{CJK}
\usepackage{multirow}
\usepackage{booktabs}
\usepackage{hyperref}

\title{Conversational Query Rewriting with Self-supervised Learning}
\name{Hang Liu, Meng Chen\sthanks{Corresponding author.}, Youzheng Wu, Xiaodong He, Bowen Zhou}
\address{JD AI, Beijing, China \\
\{\textit{liuhang55, chenmeng20, wuyouzheng1, xiaodong.he, bowen.zhou\}@jd.com}}
\begin{document}
%
\maketitle
\begin{abstract}
Context modeling plays a critical role in building multi-turn dialogue systems. Conversational Query Rewriting (CQR) aims to simplify the multi-turn dialogue modeling into a single-turn problem by explicitly rewriting the conversational query into a self-contained utterance. However, existing approaches rely on massive supervised training data, which is labor-intensive to annotate. And the detection of the omitted important information from context can be further improved. Besides, intent consistency constraint between contextual query and rewritten query is also ignored. To tackle these issues, we first propose to construct a large-scale CQR dataset automatically via self-supervised learning, which does not need human annotation. Then we introduce a novel CQR model \textbf{Teresa} based on Transformer, which is enhanced by self-attentive keywords detection and intent consistency constraint. Finally, we conduct extensive experiments on two public datasets. Experimental results demonstrate that our proposed model outperforms existing CQR baselines significantly, and also prove the effectiveness of self-supervised learning on improving the CQR performance.

\end{abstract}

\section{Introduction}
\label{sec:intro}
Building conversational bots has attracted increasing attention due to the promising potentials on applications like virtual assistants \cite{shum2018eliza} or customer service systems \cite{li2017alime}. With the development of deep learning, both the task-oriented dialogue and open-domain conversation have made remarkable progress in recent years \cite{li2017end,sordoni2015neural,chen2019sequential}. However, multi-turn dialogue modeling still remains extremely challenging. One major reason is people tend to use co-reference and ellipsis in daily conversations \cite{carbonell1983discourse}, which leaves the utterances paragmatically incomplete if they are separated from context. According to previous research, this phenomenon exists in more than 60\% conversations \cite{pan2019improving}. Taking the conversation in Table 1 for example, the key information of \textit{Bluetooth headphones} is omitted in the $Q_2$. To help the conversational bots understand the incomplete utterances, we rewrite $Q_2$ to $R_2$.

\begin{table}[t]
    \centering
    \begin{tabular}{cl}
     \hline
     \textbf{Turn} & \textbf{Utterance} (\textit{Translation})\\
     \hline
     \multirow{2}*{$Q_{1}$}  & \begin{CJK*}{UTF8}{gbsn}请问Mix3可以连接\textcolor[RGB]{255,0,0}{蓝牙耳机}吗?\end{CJK*} \\
     & \textit{Can Mix3 connect to \textcolor[RGB]{255,0,0}{Bluetooth headphones}?} \\
     \multirow{2}*{$A_{1}$}
     & \begin{CJK*}{UTF8}{gbsn}可以的\end{CJK*} \\
     & \textit{Yes, Mix3 can.} \\
     \multirow{2}*{$Q_{2}$}
     & \begin{CJK*}{UTF8}{gbsn}小米8可以连接吗?\end{CJK*} \\
     & \textit{Can Mi8 connect it?} \\
     \hline
     \multirow{2}*{$R_{2}$}
     & \begin{CJK*}{UTF8}{gbsn}小米8可以连接\textcolor[RGB]{255,0,0}{蓝牙耳机}吗?\end{CJK*} \\
     & \textit{Can Mi8 connect to  \textcolor[RGB]{255,0,0}{Bluetooth headphones}?} \\
     \hline
    \end{tabular}
    \caption{An example of contextual query rewriting. The incomplete query $Q_{2}$ is rewritten into $R_{2}$ by our proposed model. Mix3 and Mi8 are model names of cellphone.}
    \label{tab:example1}
\end{table}

Previous works \cite{pan2019improving,su2019improving,song2020two,zhang2020filling,liu2020incomplete,yu2020few} formulate the conversational context understanding as a query rewriting problem, transforming a user utterance with anaphora or ellipsis into a new utterance where the left-out or referred expressions are automatically generated from the dialogue context. Usually an end-to-end sequence-to-sequence model with copy mechanism is applied for this task. Su et al. \cite{su2019improving} proposed a Transformer-based generative model with pointer network. To locate the omitted information from context, Song et al. \cite{song2020two} employed a multi-task learning framework by taking sequence labeling as an auxiliary task. Pan et al. \cite{pan2019improving} proposed a Pick-and-Combine model to decompose the task into a cascaded process. The picking stage predicts the omitted words and the combining stage rewrites the query. As the training process of generative model needs massive rewriting pairs, all above works construct their datasets by manual annotation.

Although tremendous progress has been made, we argue that the following aspects can be further improved. First, collecting large-scale supervised data is extremely time-consuming and labor-intensive, which becomes the bottleneck of neural models. Second, the sequence labeling task tends to focus on entities and may ignore other important information, such as verb and adjective words. However, the omitted information is usually text spans which are not limited to entity words. Third, previous works lack intent consistency constraint between contextual query and rewritten query, which leaves the generation under-constrained.

To tackle above issues, in this paper, we propose a novel \textbf{T}ransformer-based qu\textbf{e}ry \textbf{re}writing model, equipped by \textbf{S}elf-Attentive Keywords Detection (SAKD) and Intent Consistency Constr\textbf{a}int (ICC), namely \textbf{Teresa}. Specifically, SAKD utilizes the self-attention weights of words to build a graph network on encoder to represent relevance between words. Then TextRank \cite{mihalcea2004textrank} algorithm is adopted to calculate each word's importance, which guides the copy mechanism during generation. As to ICC, we first obtain the intent representations of contextual query and rewritten query with the same encoder, then force their distributions on intent to keep consistent by Kullback-Leibler divergence loss \cite{kullback1997information}. Lastly, we propose to construct the CQR training data automatically from raw dialogue corpus with self-supervised learning (SSL) \cite{liu2020self}, which does not need manual annotation. Extensive experiments are performed on two public datasets. And experimental results demonstrate the superiority of our proposed model compared with state-of-the-art baselines. 


\section{Methodology}
\label{sec:approach}
We denote a conversation session $s = \{u_{1}, u_{2}, ... ,u_{t}\}$ with $t$ utterances. Given $q = u_{t}$ is the incomplete query and $c = \{u_{1},...,u_{t-1}\}$ is the context, our goal is to learn a rewriting model $g(c, q)$ to generate a context-independent query $r$, which has the same meaning with $q$ but recovers all co-referenced and omitted information. $r$ could be equivalent to $q$ when $q$ is already self-contained without context $c$. 

\subsection{Self-Supervised Learning}
\label{ssec:self-supervised learning}
Generally, the supervised learning (SL) is trained over a specific task with a large manually labeled dataset. 
Differently, self-supervised learning (SSL), also known as self-supervision, is an emerging solution to such cases where data labeling is automated, and human interaction is eliminated. In SSL, the learning model trains itself by leveraging one part of the data to predict the other part and generate labels accurately. In the end, this learning method converts an unsupervised learning problem into a supervised one. While Computer Vision is making amazing progress on SSL only in the last few years \cite{jing2020self}, SSL has been a trend in NLP research recently. Especially for the representation learning in NLP (e.g. Skip-Gram \cite{mikolov2013efficient} and BERT \cite{devlin2019bert}), various \textit{pre-training tasks} are proposed in the self-supervised formulations, such as Neighbor Word Prediction, Masked Language Modeling, and Next Sentence Prediction etc.

\begin{figure}[t]
\begin{minipage}[t]{1.0\linewidth}
  \centering
  \centerline{\includegraphics[width=9.0cm]{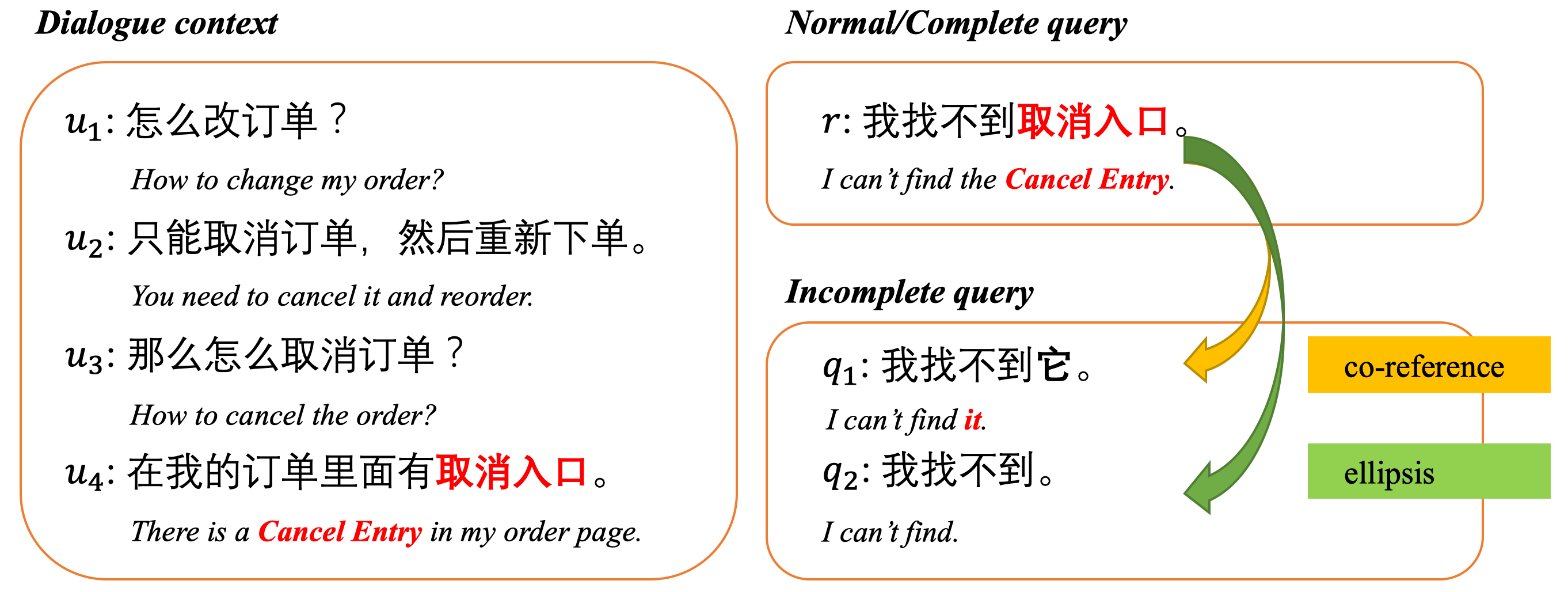}}
\end{minipage}
\caption{Dataset construction based on SSL.}
\label{fig:ssl}
\end{figure}
Inspired by SSL, as Figure 1 shows, we propose to construct the training sample ($c$, $q$, $r$) from raw dialogue corpus automatically, by corrupting the normal query $r$ into incomplete query $q$. Suppose there exist common text spans between context $c$ and normal query $r$, we can construct the incomplete $q$ by treating the common text spans by following two approaches: (1) removing the common text spans from $r$ directly, (2) if the common text spans are noun phrases in $r$, we replace them with pronouns in 50\% of time. The first approach is designed to cover the ellipsis situation, and the second approach is to cover the co-reference scenario. To improve the quality of the constructed dataset, we require the common text spans to contain at least one word of noun, verb or adjective. And only informative queries (queries with at least 10 characters for Chinese in this work) are processed. With the constructed dataset above, the CQR task can be formulated as a typical SSL problem. We first corrupt the normal query $r$, then force the model to recover it.

We also follow the trendy \textit{pre-train and fine-tuning} two-stage learning paradigm to train our CQR model. The pre-training stage is based on SSL with auto-generated data, and the fine-tuning stage is based on SL with annotated data. 

\subsection{Model}
\textbf{Transformer-based Generative Model.}
Figure 2 shows the overall architecture of our proposed model Teresa, which is in the encoder-decoder framework \cite{bahdanau2015neural}. Both the encoder and decoder are based on the transformer model \cite{vaswani2017attention}. To learn dependency between context $c$ and query $q$, the input context and query are packed together with a segment token [SEP]. For each token $w_{i}$, the input embedding is the sum of token, position and segment embedding where segment embedding indicates if each token comes from the context or query. Then transformer encoder is leveraged to produce a sequence of hidden states $H$.

\begin{figure}[htb]
\begin{minipage}[b]{1.0\linewidth}
  \centering
  \centerline{\includegraphics[width=9.5cm]{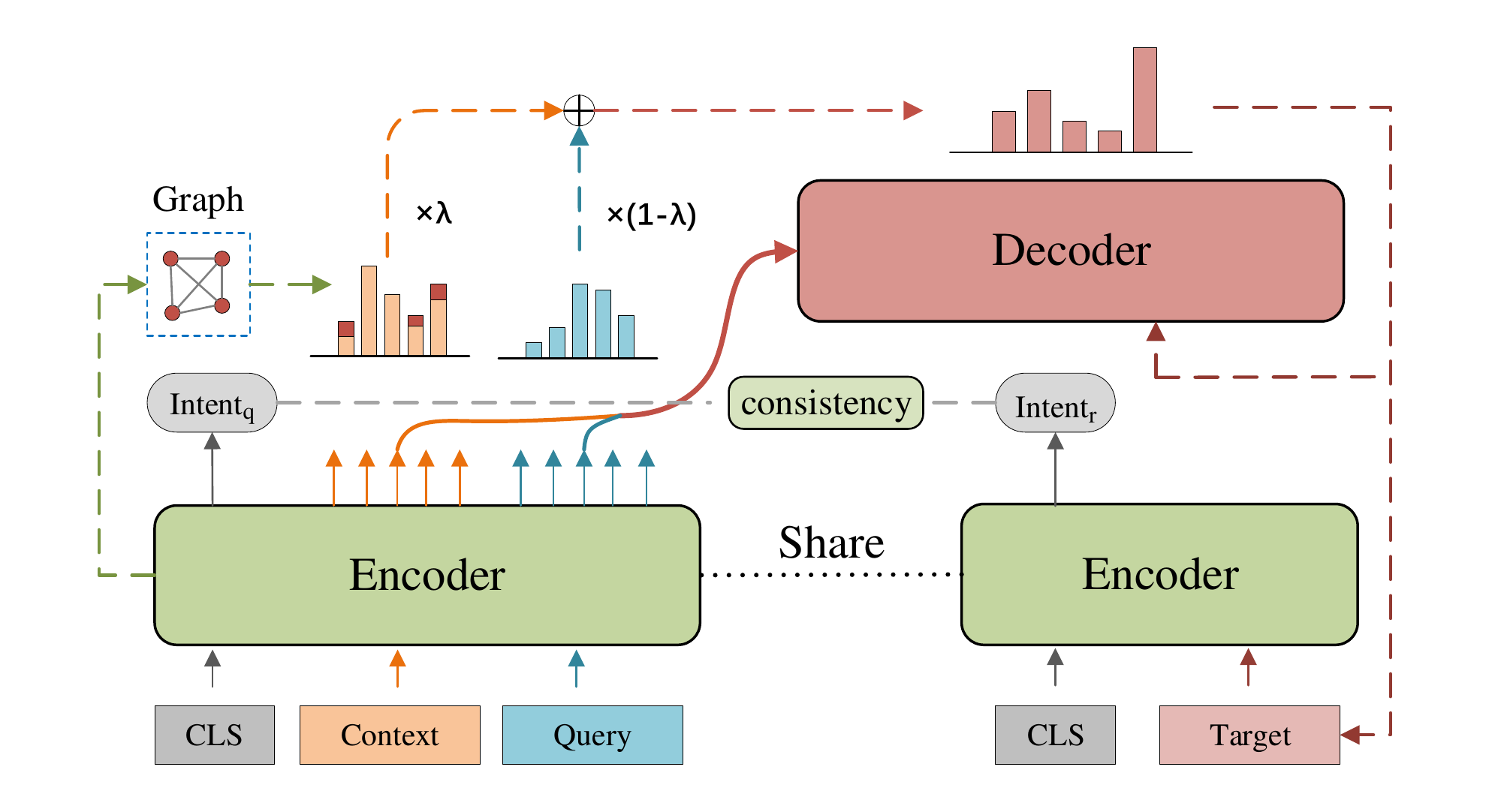}}
\end{minipage}
\caption{The framework of our propose model \textbf{Teresa}.}
\label{fig:model}
\end{figure}
The transformer decoder is applied to generate rewritten query ${r}$, in which the copy mechanism \cite{gu2016incorporating} is utilized to copy important words from the context $c$ and query $q$. Each layer $l$ of decoder is composed of three sub-layers. The first sub-layer is a multi-head self-attention layer $M^{l}$. The second sub-layer is an encoder-decoder interaction layer. And the third sub-layer is a feed-forward layer. Inspired by Su et al \cite{su2019improving}, we calculate the context representation $C^{l}$ and query representation $Q^{l}$ separately in the encoder-decoder interaction sub-layer, and $C^{l}$ and $Q^{l}$ are concatenated as input of the feed-forward sub-layer to obtain the final decoder hidden state $S^{l}$. At each time step $t$, the decoding probability $P(w)$ is computed by fusing the information from $c, q$ and last decoding layer hidden state $S_{t}$. The copy mechanism is used to predict the next target word according to $P(w)$, which is computed as follows:
\begin{equation}
P(r_{t}=w)=\lambda\sum_{i:(w_{i}=w,\in c)} a_{t,i}+(1-\lambda)\sum_{j:(w_{j}=w,\in q)}a'_{t,j} 
\end{equation}
\begin{equation}
    a_{t} = Attention(M_{t}, H_{c})
\end{equation}
\begin{equation}
    a'_{t} = Attention(M_{t}, H_{q})
\end{equation}
\begin{equation}
    \lambda = \sigma(w_{S}^{T}S_{t} + w_{C}^{T}C_{t} + w_{Q}^{T}Q_{t})
\end{equation}
where $w_{S}$, $w_{C}$, and $w_{Q}$ are trainable parameters. $\sigma$ is the sigmoid function to output a value between 0 and 1. $\lambda$ is a learning coefficient to decide whether to copy token from $c$ or $q$. Note that all tokens in $r$ can only be copied either from the context $c$ or query $q$. The rewriting model is trained by maximizing the log-likelihood of the output rewritten query.
\begin{equation}
    \mathcal{L}_{NLL} = -\frac{1}{T} \sum_{t=0}^{T} logP(r_{t})
\end{equation}

\textbf{Self-Attentive Keywords Detection.}
To facilitate carrying the omitted important information from context into the rewritten query, we propose to enhance the copy mechanism by a novel Self-Attentive Keywords Detection module (SAKD). Inspired by Xu et al \cite{xu2020self}, we build a graph network by stacking a self-attention layer over the output of context encoder. Formally, let $G=(V,D)$ be a directed graph where vertices $V$ is word set from context and edge $D_{i,j}$ represents the self-attention weight from word $w_{i}$ to the word $w_{j}$. Then the TextRank algorithm \cite{mihalcea2004textrank} is adopted to calculate word importance $score$ based on graph $G$. 

The word importance score can be seen as a prior information to indicate the salient information in dialogue context. It is incorporated into copy mechanism as an extra input to calculate the attention weights $a_{t}$ of context words. To further ensure that important information is extracted by copy mechanism, the Kullback-Leibler (KL) divergence is adopt as an auxiliary loss to force the distribution of attention weights close to the prior importance $score$. 
\begin{equation}
    a_{t} = softmax(Attention(M_{t}, H_{c}) + w_{score}^{T}score)
\end{equation}
\begin{equation}
    \mathcal{L}_{SAKD} = KL(\frac{1}{T}\sum_{t=0}^{T}a_{t}, score)
\end{equation}

\textbf{Intent Consistency Constraint.}
Intent matters to query understanding in dialogue. We argue that the rewritten query should be consistent with the contextual query in the intent dimension. Therefore, we propose a novel Intent Consistency Constraint (ICC) to guide the rewriting process. In this module, the latent intent recognition task is equipped to learn the corresponding intent representation for the given contextual query. The latent intent recognition shares encoder parameters with the rewriting model. A special classification token [CLS] is inserted in front of the input sequence to collect the intent information of original query in the context. Similarly, the corresponding intent representation for the rewritten query can also be collected by the content of itself, because the rewritten query is self-contained. Then another KL divergence loss is adopted to keep the intent distributions consistency between the contextual query and the rewritten query. 

To sum up, the total objective of our proposed model is to minimize the integrated loss:
\begin{equation}
    \mathcal{L} = \mathcal{L}_{NLL} + \mathcal{L}_{SAKD} + KL(f(H_{CLS}^{c,q}), f(H_{CLS}^{r}))
\end{equation}
where $f$ is a function to map text representations into intent distributions.
 
\section{Experiments}
\label{sec:experiments}
\subsection{Datasets and Metrics}
\label{ssec:dataset}
We carry out extensive experiments on two public datasets. First, we construct a new CQR dataset from scratch based on large-scale raw dialogue corpus JDDC \cite{chen2020jddc}. The JDDC corpus contains more than 1 million real multi-turn conversations between users and customer service staffs in E-commerce scenario. The average turn number of dialogues is 20, indicating the contextual dependency is very common in the dialogues. By applying the SSL approach mentioned in Section 2.1, we generate the pre-training data \textit{JDDC-CQR-10M}, which includes about 10 million \textit{(c, q, r)} triplets. We generate the positive samples by SSL, and the negative samples by random sampling. To compare with the SL approach, we also annotate another 146,000 triplets manually, namely \textit{JDDC-CQR-146K}. The ratio of positive and negative samples is 1:1 in above two datasets. The positive sample means $r$ is different from $q$, and the negative sample means $r$ is the same as $q$. For context $c$, we keep at most 5 utterances. Second, to compare with previous CQR models, we also conduct experiments on a public CQR dataset \textit{Restoration-200K}, which was collected from open-domain conversations and manually annotated by \cite{pan2019improving}. We split both the \textit{JDDC-CQR-146K} and \textit{Restoration-200K} into train/dev/test sets. The \textit{JDDC-CQR-10M} is only used for pre-training in our experiment.         

For evaluation, we choose three automatic evaluation metrics of BLEU-4 \cite{papineni2002bleu}, ROUGE-L \cite{lin-2004-rouge}, and Exact Match by following previous works \cite{su2019improving}. BLEU and ROUGE are widely used in generation tasks to measure the lexical similarity between generated utterance and ground-truth. Exact Match is a very strict metric which requires the generated utterance to be the same as the ground-truth. 

\begin{table}[t]
    \centering
    \begin{tabular}{lcccc}
     \hline
     \hline
     \textbf{Model} & \textbf{B4} & \textbf{RG-L} & \textbf{EM(-)} & \textbf{EM(+)} \\
     \hline
     \textit{JDDC-CQR-146K} & \\
     T-Ptr-$\lambda$ \cite{su2019improving} & 73.34 & 84.71 & 96.55 & 32.40 \\
     PAC \cite{pan2019improving} & 70.78& 83.52& 87.18& 28.24 \\
     MLR \cite{song2020two} & 68.53 & 81.12 & 92.65 & 19.08 \\
     \hline
     Teresa w/ SL & 73.71 & 84.90 & 96.60 & 33.07 \\
     Teresa w/ SSL & 78.34 & 87.68 & 94.94 & 47.36\\
     Teresa w/ SSL+SL & \textbf{79.62} & \textbf{88.78} & \textbf{97.59} & \textbf{50.82}\\
     \hline
     w/o SAKD & 79.35 & 88.61 & 97.36 & 50.00 \\
     w/o ICC & 78.81 & 88.25 & 97.07 & 48.69 \\
    \midrule[0.8pt]
    \textit{Restoration-200K} & \\
    T-Ptr-$\lambda$ \cite{su2019improving} & 74.73 & 88.65 & 86.63 & 53.63 \\
    PAC \cite{pan2019improving} & 73.69 & 86.66 & 82.23 & 46.27 \\
    MLR \cite{song2020two} & 71.99 & 86.74 & 82.42 & 48.01 \\
    \hline
    Teresa w/ SL & \textbf{74.82} & \textbf{88.69} & \textbf{87.49} & \textbf{54.46} \\
    \hline
    \hline
    \end{tabular}
    \caption{The experimental results on \textit{JDDC-CQR-146K} and \textit{Restoration-200K} datasets. \textbf{B4} and \textbf{RG-L} stand for BLEU-4 and ROUGE-L respectively. \textbf{EM(+)} and \textbf{EM(-)} represent EM percentage for positive and negative samples.}
    \label{tab:experimental result}
\end{table}

\subsection{Baselines}
\label{ssec:baselines}
Our baselines are as follows: (1) \textbf{T-Ptr-$\lambda$} \cite{su2019improving}. This is a Pointer-Generator Network based on Transformer, which only copies words from context or query during generation. (2) \textbf{PAC} \cite{pan2019improving}. Pick-and-Combine (PAC) is a cascaded model that first identifies omitted words in context based on BERT \cite{devlin2019bert}, then appends the omitted words to the query as the input of Pointer-Generator Network. (3) \textbf{MLR} \cite{song2020two}. MLR is a two-stage CQR model with multi-task learning, which trains the sequence labeling task and query rewriting task jointly. All three baseline models are trained with the annotated training set (\textit{aka. supervised learning}) in the following experiments. 

For our proposed model \textbf{Teresa}, both the encoder and decoder consist of 6 layers of transformer block. The embedding dimension is set to 256 and the attention head number is 8. It is optimized with the Adam optimizer. The initial learning rate is 0.5 and batch-size is 64. Beam search is used for decoding and beam size is 4. For all the baselines, we follow the same experimental settings in the corresponding papers. 

\subsection{Experimental Results}
\label{ssec:experiment}
Table 2 shows the experimental results on \textit{JDDC-CQR-146K} and \textit{Restoration-200K}. We can obtain following interesting conclusions: (1) For both two CQR datasets, with only the annotated training data, Teresa w/ SL outperforms above three baselines on all metrics. It indicates the superiority of our proposed model. (2) Even with the pre-training data only, Terera w/ SSL has already outperformed Terera w/ SL significantly on \textbf{B4}, \textbf{RG-L} and \textbf{EM(+)}, which proves the effectiveness of SSL. For \textbf{EM(-)}, we argue it may be slightly effected by the random negative samples. (3) By utilizing the \textit{pre-train and fine-tune} paradigm, Teresa w/ SSL+SL makes further improvement and obtains the best performance. 

To figure out the contributions of SAKD and ICC, we conduct two groups of ablation study on \textit{JDDC-CQR-146K}. From Table 2, it's observed that, by removing SAKD and ICC separately, both the performance drops notably, which demonstrates the necessity and rationality of each module. 

Figure 3 illustrates the performance when we fine-tune Teresa with different percentages of annotated data. The two curves show that the performance improves very fast when adding only 10\% of annotated data. Then it starts to saturate even adding more annotated data. This indicates much less of annotated data is needed with the help of SSL. We tried to plug our CQR model in the dialogue system and the experiments show that CQR can facilitate downstream tasks too.

\begin{figure}[t]
  \centering
  \centerline{\includegraphics[width=9.0cm]{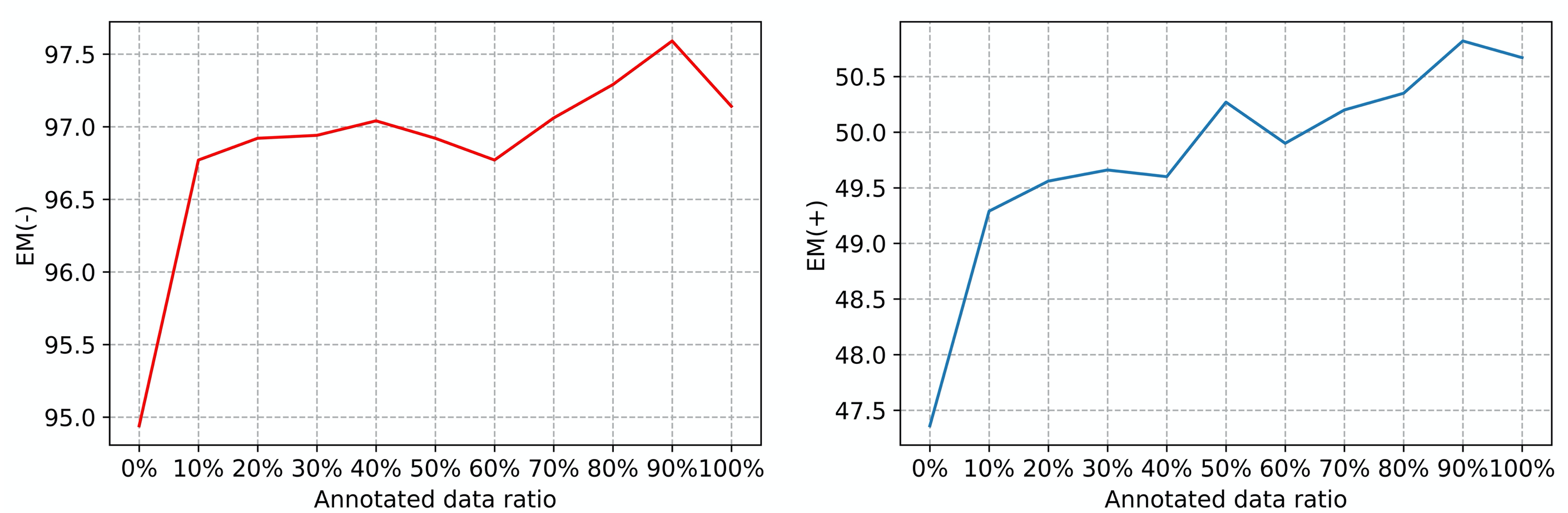}}
\caption{Performance analysis of fine-tuning experiments.}
\label{fig:res}
\end{figure}

\section{Conclusions}
\label{sec:conclusions}
In this paper, we propose a novel transformer-based generative model (denoted as Teresa) for conversational query rewriting, which is equipped by a novel self-attentive keywords detection module and an auxiliary intent consistency constraint. To address the time-consuming data annotation issue, we propose to construct the CQR training data via self-supervised learning automatically. Experiments on two CQR datasets demonstrate the superiority of SSL and the competitiveness of our proposed model. In the future, we will explore integrating the CQR task into pre-training stage of Pre-trained Language Model, and provide an universal pre-trained CQR model for various dialogue tasks.    

\section{Acknowledgement}
\label{sec:acknowledgement}
This work was supported by the National Key R\&D Program of China under Grant No.2020AAA0108600. Thanks to Shaozu Yuan for technical discussions.



\vfill\pagebreak

\bibliographystyle{IEEEbib}
\bibliography{strings,refs}

\end{document}